# Knee Osteoarthritis Severity Grading Using Optimized Deep Learning and LLM-Driven Intelligent AI on Computationally Limited Systems

Dayam Nadeem*
*Department of Computer Science & Engineering*
*Jamia Hamdard*
New Delhi ,India
dayam8696@gmail.com

Neha
*Department of Computer Science & Engineering*
*Jamia Hamdard*
New Delhi ,India
nehanit2807@gmail.com

Safdar Mustafa
*Department of Computer Science & Engineering*
*Jamia Hamdard*
New Delhi ,India
safdarmustafa01@gmail.com

Adnan Alvi
*Department of Computer Science & Engineering*
*Jamia Hamdard*
New Delhi ,India
adnanalvi856@gmail.com

Mohd Hussain
*Department of Computer Science & Engineering*
*Jamia Hamdard*
New Delhi ,India
md28md.hussain@gmail.com

***Abstract*—Knee osteoarthritis (KOA) is among the musculoskeletal disorders that considerably restrict joint mobility, cause severe chronic pain and impact negatively on quality life. It is one of the persistent health issues worldwide. Generally, subjectivity and inter-observer variability undermine conventional practices and evaluation process that are adopted to address such health issues. Hence precise and timely diagnosis would be one of the effective ways for the assessment of its severity. This paper proposes an automated diagnostic approach for severity grading of KOA by blending a deep learning convolutional neural network (CNN) with a device-based inference platform powered by TensorFlow Lite. It proposes a model based on the ResNet-18 convolutional neural network. The designed model is trained on publicly available database. Through a transfer learning approach obtained knee images are first classified into five Kellgren–Lawrence (KL) grades. Further the developed model is optimised. During the training of the model test accuracy of 94.48%. with stable convergence has been achieved. Subsequently the optimised model transformed into a lightweight TensorFlow Lite format, facilitating seamless deployment on resource-constrained devices. The designed model is capable enough to operate in the environment having no continuous internet connectivity. Also, an auxiliary Large Language Model (Gemini-2.0-flash) is applied to generate structured interpretive findings like potential symptoms, risk factors, and preventive majors etc. The LLM component functions as interface without influencing the classification process. The proposed model articulates the feasibility of an on-device, interpretable decision-support tools for early diagnosis and improve accessibility to Artificial Intelligence (AI)-assisted knee screening tool.***

## I. Introduction

Knee osteoarthritis (KOA) is a joint disease that causes articular cartilage to progressively deteriorate, osteophytes to develop, and subchondral bone to undergo structural transformations. Globally, KOA affects millions of individuals who experience chronic pain and reduced mobility, which results in decreased quality of life. The knee joint represents one of the most frequently affected areas, as over 250 million people worldwide deal with osteoarthritis [1]. The condition becomes more common among individuals who have reached an older age and those who weigh more than normal and people who have a genetic family history and individuals who have suffered from previous joint injuries. The condition imposes serious economic costs because it requires continuous medical treatment, which results in lost work time and prevents people from doing their regular activities. The Kellgren-Lawrence (KL) grading system provides an X-ray assessment tool that medical professionals use to evaluate the severity of KOA through radiographic imaging[2]. The grading system divides radiographic progression into five grades according to joint space gap and osteophyte formation [3]. Early detection and precise assessment of conditions continue to be difficult because manual interpretation is hampered by subjective assessment and different observers evaluating the same content at different levels. The system has these limits, which make it impossible to use for routine assessments of KOA. The establishment of an automated assessment system has become essential because it will assist common people and medical professionals with their diagnosis and clinical decision-making procedures.

Latest developments in artificial intelligence (AI), especially deep learning models such as convolutional neural networks (CNNs)-which are AI models designed to automatically learn visual patterns directly from images, and ResNets, an enhanced form of CNN that enables training of deeper networks using skip connections, have shown strong potential in automating KOA grading and aiming to assist healthcare professionals in decision-making. The process of automated classification together with LLM interpretive insights enables common people and healthcare professionals to assess knee health conditions through fast and dependable methods that deliver clinically relevant study results. LLMs (large language models) are AI systems trained on large text datasets that generate human-readable explanations from structured outputs, and they have grown into helpful resources for generating easily interpretable clinical analysis from structured model outputs. The combination of LLM-based interpretations with image-based deep learning systems improves usability and accessibility while enhancing understanding and enabling decision support systems to maintain their core prediction accuracy.

The combination of growing elderly populations and increasing rates of obesity has led to higher cases of KOA which needs both preventive and early detection solutions for public use which this AI model provides through its on-device implementation.

The research presents an improved system that automatically evaluates knee osteoarthritis through its dedicated assessment model. The proposed model is deployed by combining a fine-tuned ResNet-18 classifier with a smart device-deployable inference engine and LLM-assisted interpretive module. The solution operates through TensorFlow Lite, which serves as a mobile-friendly deep learning system to perform efficient on-device processing, which can function without internet access on devices with restricted processing capabilities. The model provides direct use for both point-of-care and remote medical settings. The LLM component presents organized details about risk factors, symptoms, preventive majors and lifestyle recommendations which enhance user experience while maintaining clinical importance through a more structured approach.

The article proceeds to present a complete examination of all advancements which took place during the process of assessing the suggested model. Section II of this paper analyzes previous studies which examined deep learning methods for KOA identification and severity assessment together with their application to current technological advancements. The exploratory data analysis and dataset characteristics of the study are presented in Section III, while Section IV explains the research approach, which includes pre-processing steps and model development processes together with training and execution methods. The following section presents results which together with evaluation results, show how well the proposed model performs. The study evaluates how well the model performs on the clinically challenging grades, which include KL 1-3 assessment. The concluding section presents a summary of all major study findings and research contributions.

## II. RELATED WORK

An article authored by Rani et al. used an advanced CNN for knee osteoarthritis detection, achieving 92.3% accuracy in binary and 78.4% in multi-class classification from X-ray images. The model achieved its best performance with binary F1 results at 96.5% and recall results at 97.05%, while its multi-class F1 results produced scores of 0.87, 0.74, and 0.67 for healthy, moderate, and severe cases. They recommend incorporating MRI and multi-source data to improve accuracy [4]. In a 2023 study, Mohammed et al. employed residual neural networks for osteoarthritis detection and severity classification on preprocessed knee X-rays. By integrating weighted loss with SMOTE, severe KOA detection rates were raised by 18–22%, allowing medical teams to identify high-risk patients earlier [5]. Sonobe et al. (2024) found that depression and severe knee pain were the primary factors that reduced physical function, while radiographic KL grade 4 was the only point where KOA severity was linked with physical impairment. The researchers advise medical professionals to evaluate radiographic severity in addition to psychological and pain factors. [6]. Tiulpin et al. (2020) created a deep learning technique that predicted KL and OARSI grades from knee x-rays using 50-layer residual networks. The system outperformed current OA detection techniques, achieving κ results of 0.94 and AUC results of 0.98. [7]. According to Neogi et al. (2013), the progression of knee osteoarthritis from intermittent to chronic pain is caused by a number of factors that call for a deeper mechanistic understanding in order to develop effective treatments that lower the costs of osteoarthritis to society [8]. Tariq et al. (2023) described KOA as a progressive disease that medical professionals can identify through radiographic assessment, which requires them to detect the condition during its early stages in order to prevent further joint deterioration and to understand how the disease affects knee function by examining its symptoms and risk factors [9]. Ahmed et al. (2024) applied explainable artificial intelligence through a divide-and-conquer method to classify KOA disabilities. The model reached 99.13% accuracy when it classified normal and severe cases. The system showed a performance drop to 67% for intermediate grades which demonstrated its inability to match expert diagnosis [10]. Jain et al. (2024) emphasized that early KOA diagnosis prevents joint damage because aging, obesity, genetics, and injury present risks while X-rays show disease severity and MRI detects bone marrow lesions which lead to disease progression[11]. Telagam et al. (2025) used CNN models in conjunction with KL grading to classify KOA severity into five categories: doubtful, mild, moderate, severe, and standard. Compared to conventional methods that rely on patient symptoms and X-ray results, this approach allows medical professionals to identify illnesses at an earlier stage and create more effective treatment plans. [12].

## III. EXPLORATORY DATA ANALYSIS

Since the distribution and characteristics of the training set KOA dataset needed to be analyzed before any model development work could start, an exploratory data analysis was necessary. Three factors are identified by the analysis process: variations in image quality and class distribution, as well as important radiographic features that influence the performance of models. All of the key conclusions that the researchers came to after analyzing the dataset are presented in the following subsections.

### A. Sample Images

In order to visually compare healthy knee joints with various stages of osteoarthritis, the researchers evaluated X-ray images, which represent the entire range of KL grades from 0 to 4. As illustrated in Figure 1, the sample images display various radiographic features, such as joint-space narrowing, osteophyte formation, and bone deformity, to show that each KL grade represents a distinct stage of degeneration.

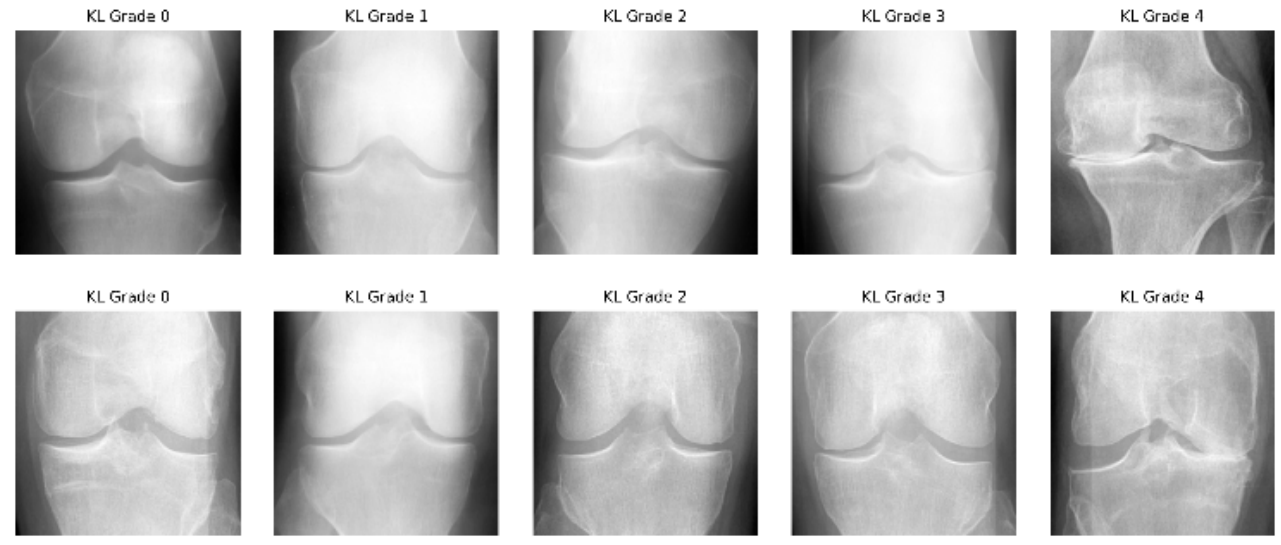

*Fig. 1 Knee X-ray dataset representative samples (KL Grades 0-4).*

This visual examination supports the comprehension of severity progression and validates the dataset's applicability.

## B. *Class Distribution*

The data distribution across Kellgren-Lawrence (KL) grades reveals a noticeable class imbalance, which results in fewer instances of severe cases. As illustrated in Figure. 2. The dataset, which consists of 5778 training images and 1656 testing images, enables researchers to study a total of 7434 knee X-ray images, which were included in this research.

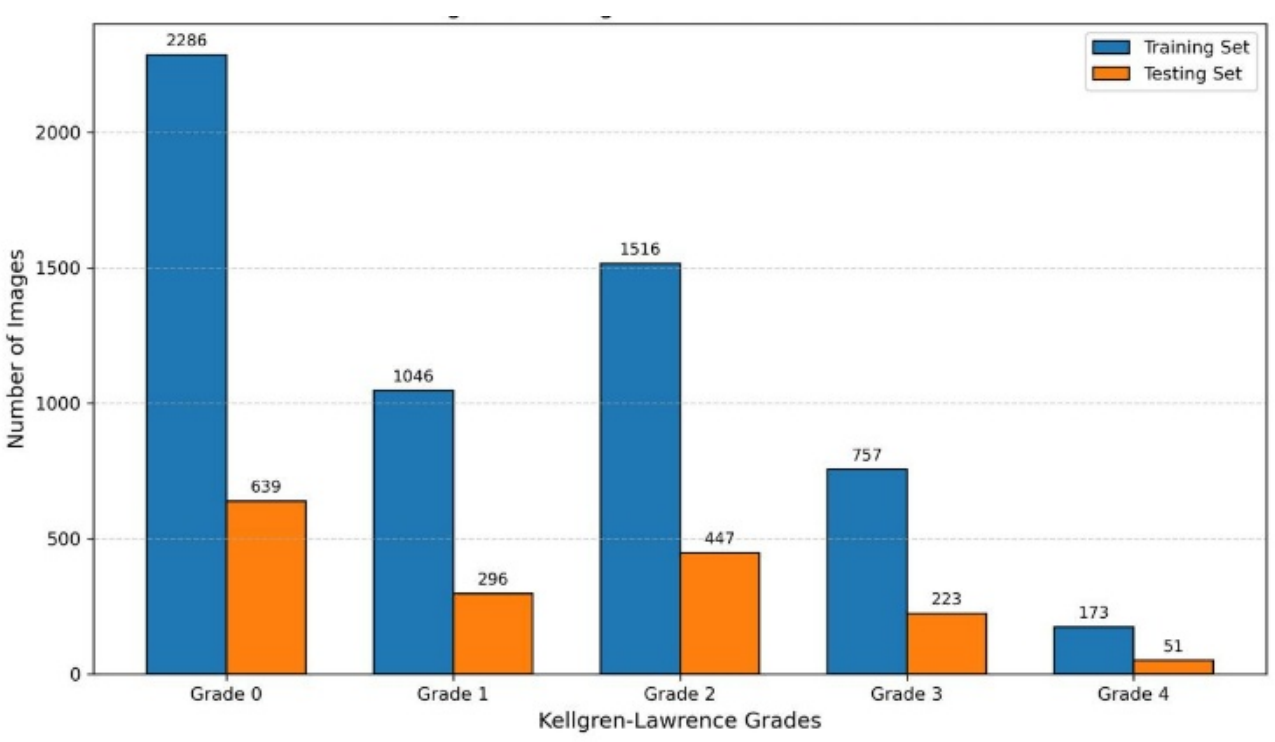

*Fig. 2 Training and testing data distribution for KL severity grades.*

The actual existence of extreme disease cases leads to the creation of unbalanced medical datasets which scientists use for research. The researchers need to identify this skew because it affects how their models learn, and it requires them to use augmentation methods together with precise training techniques to prevent their systems from favoring larger groups.

## C. *Pixel Intensity Histograms*

The creation of pixel intensity histograms for images from all five KL grades enables researchers to examine brightness and contrast patterns, which are displayed in Figure. 3. The research findings demonstrate that different severity levels produce distinct grayscale patterns because of variations in radiographic techniques and exposure settings. The study demonstrates that normalization requires standardized preprocessing methods to guarantee model input remains unchanged when intensity levels differ from one another.

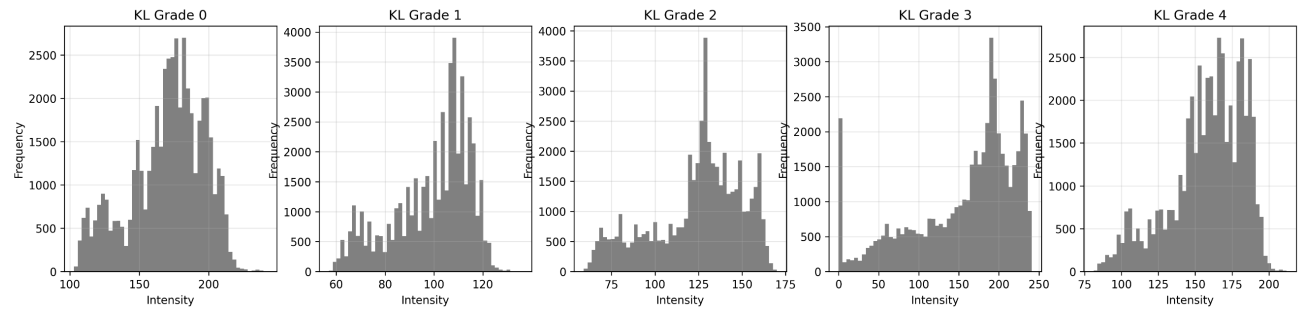

*Fig. 3 Pixel intensity histograms for knee X-ray images across KL Grades 0–4.*

# IV. Proposed Methodology

## A. *Data Pre-Processing*

The data preprocessing pipeline ensures that all X-ray inputs undergo both standardization and optimization processes required for deep learning. As illustrated in Figure. 2, The dataset divides into training and testing subsets according to all five KL severity grades, which demonstrate class imbalance and require augmentation strategies to address this issue. The dataset is divided into image segments, which follow the standard practice used for this dataset, and this method allows researchers to use their available radiographs for building and testing models. The framework will undergo expansion in future research through the addition of patient-based partitioning, which will improve evaluation accuracy and clinical strength.

Figure 4 shows the entire preprocessing process, which starts from loading the dataset and continues until the creation of batches. The process starts with the creation of two dataframes, which store image paths together with their corresponding severity labels. The entire set of images undergoes resizing to dimensions of 224×224 pixels which enables compatibility with the input specifications needed by the ResNet architecture. Training images undergo controlled augmentations, which include horizontal flipping, to create different training conditions that help prevent overfitting, while the implementation of ImageNet mean and standard deviation values for normalization maintains uniform pixel brightness across all samples in accordance with previous EDA findings. The custom PyTorch KneeDataset class handles the entire process, which includes image retrieval, RGB conversion, and transformation application, followed by the creation of batches that undergo shuffling through DataLoaders. The methodical preprocessing procedure guarantees that the model receives data that has been cleaned and arranged into suitable, balanced categories, which results in increased KOA severity classification accuracy and dependability.

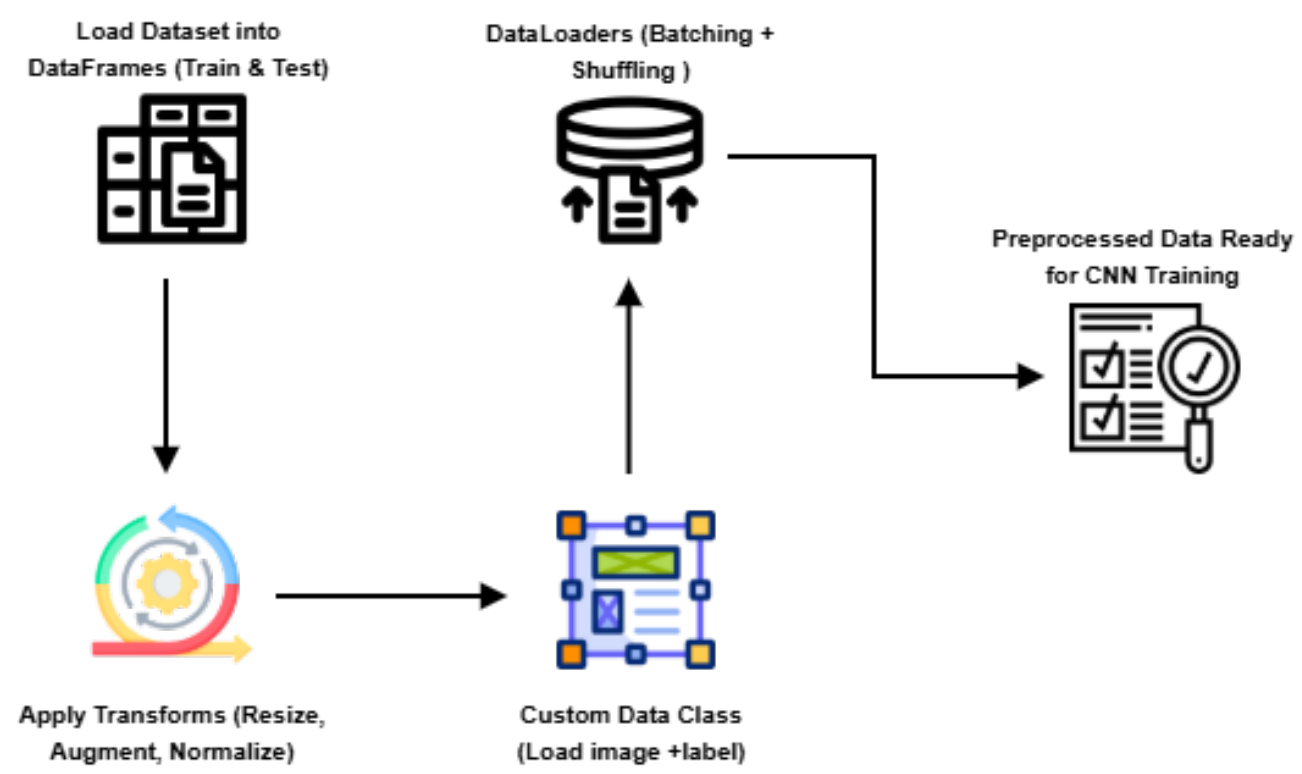

*Fig. 4 Pre-processing pipeline for knee X-ray dataset.*

## B. Model Development

The model development process focuses on creating an effective deep learning system that automatically determines knee osteoarthritis severity through the five KL grading system. The model uses transfer learning to acquire visual understanding from existing models while learning to identify specific radiographic features of the dataset. For this purpose, researchers selected a pretrained ResNet-18 system because its residual learning design helps maintain gradient movement while producing useful results from complex deep learning systems.

To create a model which can perform five-class KOA grading, the last fully connected layer of ResNet-18 needs replacement with a new linear classifier that has five output neurons, which represent KL Grades 0 to 4. The modified classification layer is expressed as

$$f_{\text{output}} = Wx + b,$$

where $W \in \mathbb{R}^{5\times d}$ and $d$ denotes the dimension of the extracted feature vector. While adjusting the final layers to radiographic severity characteristics, the network can use generic texture and structural cues learned from large-scale natural image datasets by keeping the pretrained weights in the earlier layers.

The model parameters are optimized using the Adam optimizer with a learning rate of $1 \times 10^{-4}$and a weight decay of $1 \times 10^{-4}$, which imposes a penalty on excessively large weights, thereby aiding in the prevention of overfitting. The multiclass cross-entropy loss serves as the foundation for establishing the training objective and is expressed as

$$\mathcal{L} = -\sum_{i=1}^{C} y_i \log(\hat{y}_i),$$

where $C = 5$ represents the number of output classes, $y_i$denotes the true label, and $\hat{y}_i$ is the predicted probability derived via Softmax. Adam improves the trainable parameters according to

$$\theta_{t+1} = \theta_t - \alpha \frac{\hat{m}_t}{\sqrt{\hat{v}_t} + \epsilon},$$

where $\hat{m}_t$and $\hat{v}_t$are bias-corrected moment estimates and $\alpha$ represents the learning rate.

The network is trained for 25 epochs with batches of 32 radiographs. The network processes each batch by forwarding images through the system, calculating loss, and using backward gradient propagation to modify model parameters. The model's stability during training was tested through continuous measurement of loss and accuracy throughout each training epoch. The complete workflow from input preprocessing to classification output is visually summarized in Figure. 5, showing how the fine-tuned ResNet-18 processes knee radiographs to predict the corresponding KL severity grade.

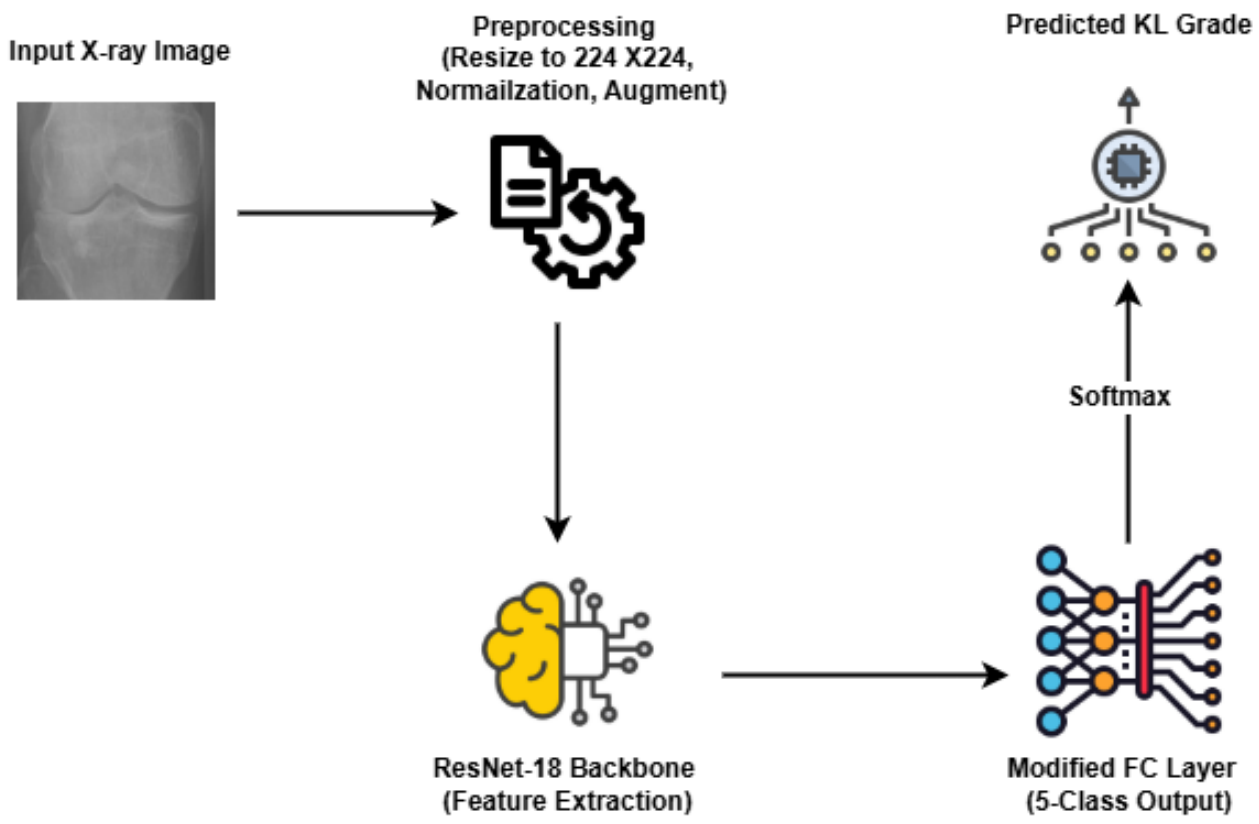

*Fig. 5 The architecture and training pipeline of ResNet-18 for KL grading.*

## C. Results and Experimental Evaluation

The proposed model for classifying KOA severity, based on ResNet-18, evaluates its operational efficiency and predictive accuracy while operating within offline resource limitations. After 25 epochs, the model illustrated in Figure. 6 attains a testing accuracy of 94.48% and a final loss of 0.1462, reflecting consistent model performance and effective feature extraction. As accuracy improves with every training session and loss consistently decreases over time, the system demonstrates robust generalization capabilities.

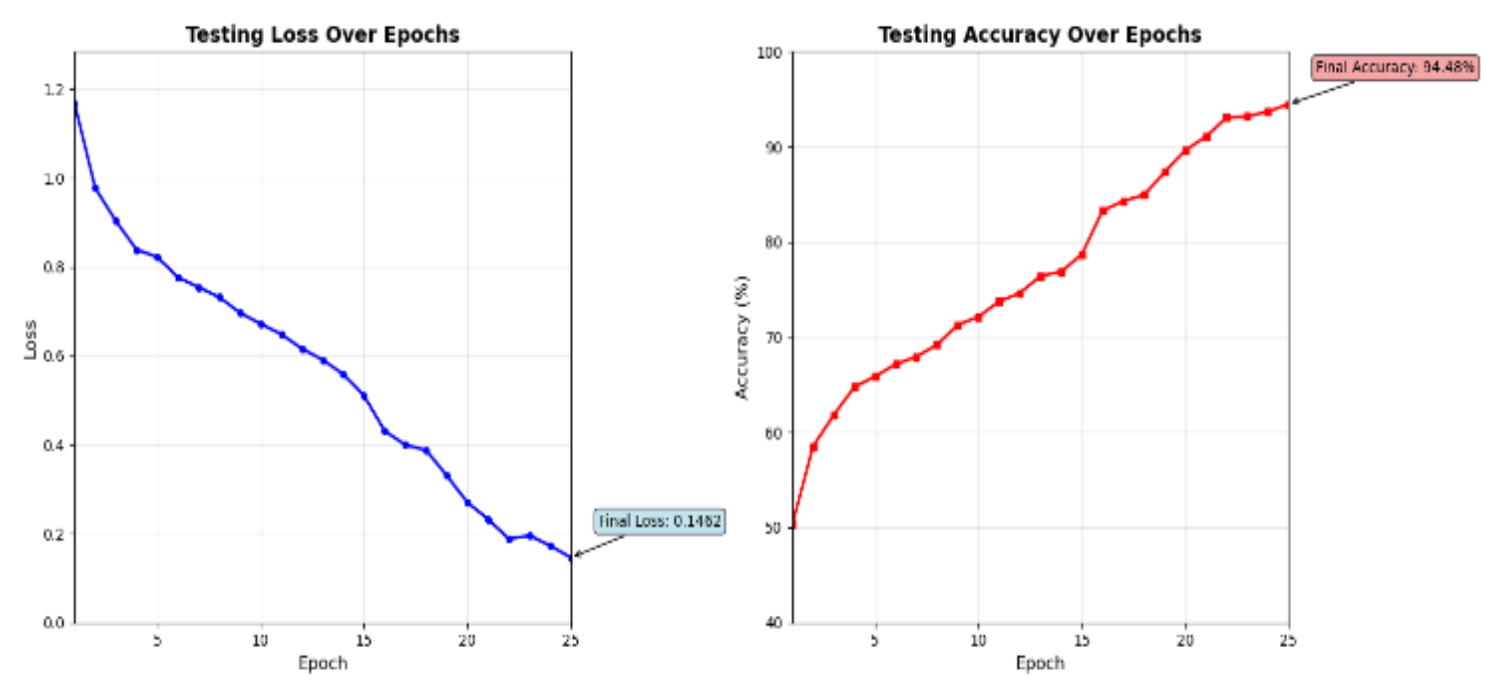

*Fig. 6 Accuracy and Loss Graph.*

To assess the real-time performance of the optimized predictive framework on devices with constrained resources, it is additionally executed in an offline environment. The deployed Android interface processes knee X-ray images through its efficient system while delivering instant KL-grade predictions that function without network connectivity, according to evidence presented in Figure. 7. The demonstrated output indicates a Grade 4: severe osteoarthritis prediction, showcasing the model's effectiveness in offline, real-time, and resource-constrained operating conditions.

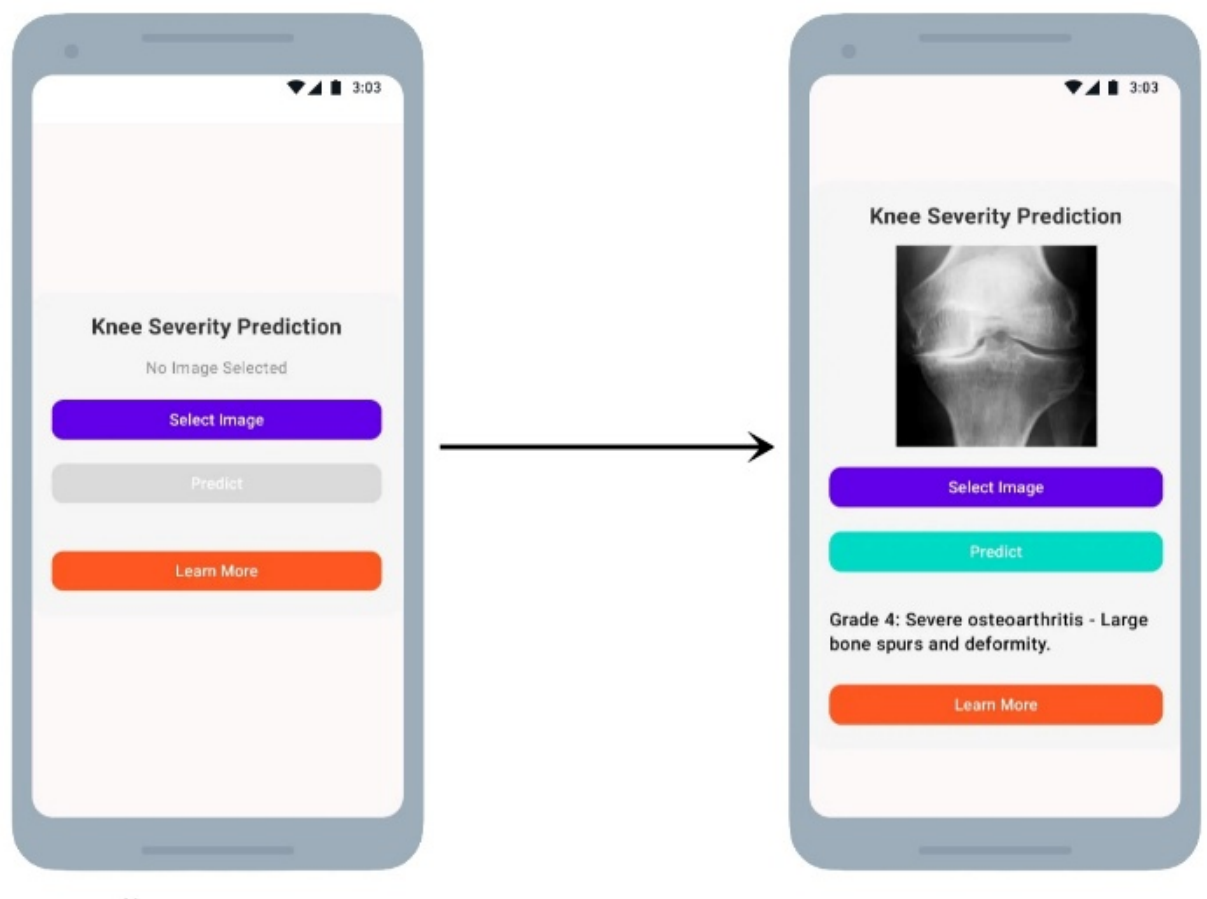

*Fig. 7 Real-time Offline Prediction Interface on Resource-Constrained Environment.*

### D. Performance Analysis on Intermediate KL Grades (1–3).

The researchers tested their framework against medical cases, which posed the biggest challenges, in order to further evaluate its strength. Automated classification systems face challenges in real-world medical settings because the intermediate KL grades, which range from 1 to 3, show less radiographic changes and more variation between observers.

The researchers combined samples from both the training and testing partitions, resulting in 4,255 images that matched KL grades 1 to 3. The confusion matrix shown in Figure.8 proves that the model maintains strong ability to differentiate between various levels of severity because it achieved its highest accurate predictions for KL grade 2.

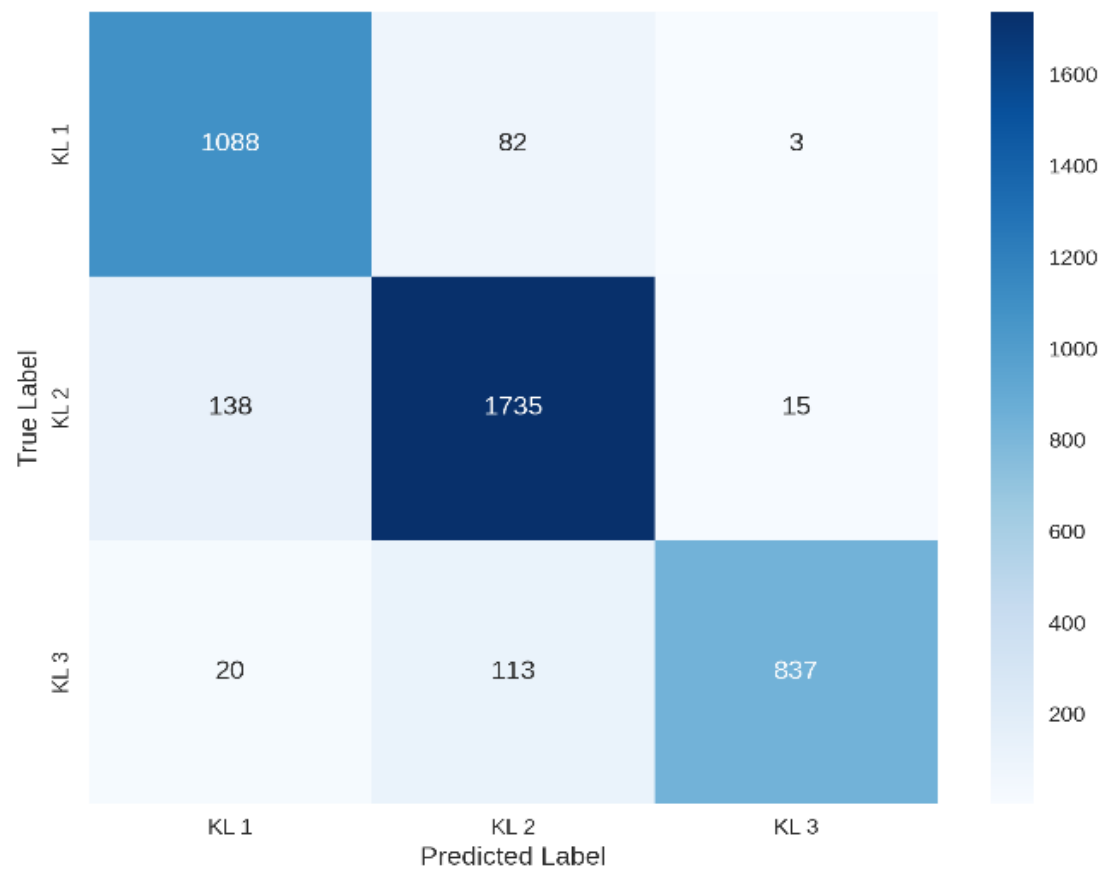

*Fig. 8 Confusion Matrix Performance on KL Grade 1-3.*

The quantitative performance metrics shown in Figure.9 demonstrate continuous performance across all three intermediate grades because the system achieves high precision and recall and F1-score results. The system demonstrates most misclassifications between adjacent grades (KL 1 vs. KL 2 and KL 2 vs. KL 3) because the gradual structural progression of knee osteoarthritis leads to this pattern, which clinicians expect.

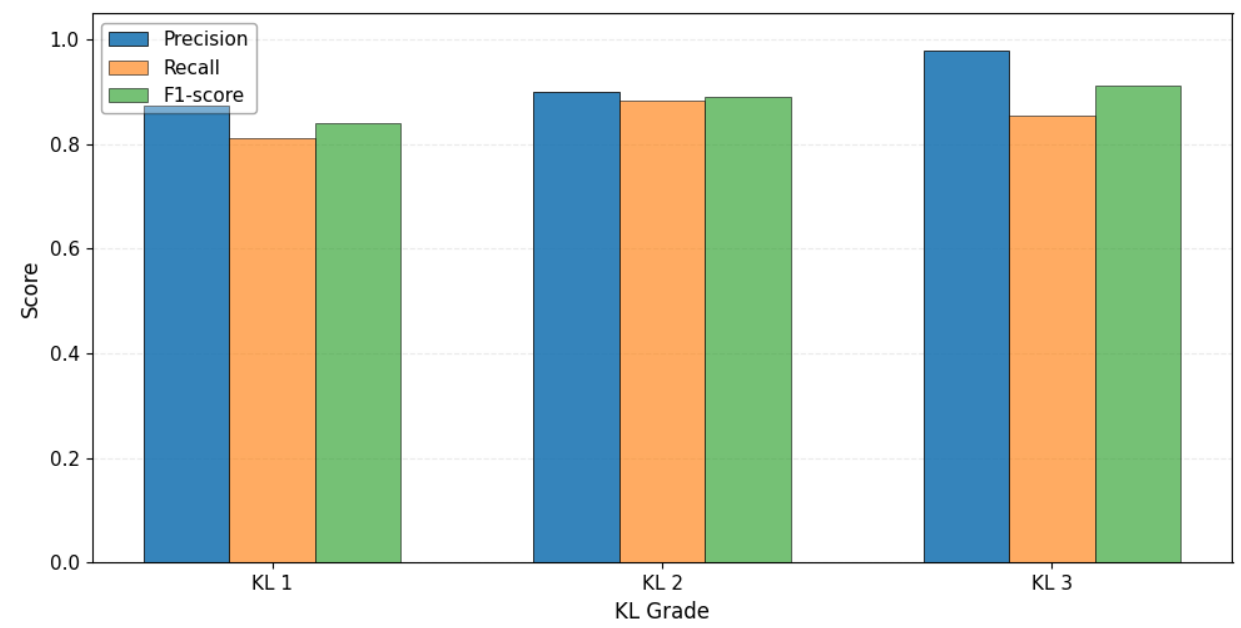

*Fig. 9 Model Performance Across Intermediate KL Grades (1-3).*

The study results show that our developed model demonstrates consistent performance in testing between two different levels of disease progression, which establishes its value for use in real-world clinical assessment and medical decision-making systems.

### E. Leveraging LLM for Interpretive Insights

#### 1) LLM Interaction Process

Once the severity grade has been predicted by the machine learning model, the result gets processed using a Large Language Model (LLM), specifically the Gemini-2.0-flash API. The predicted grade, along with a pre-defined hidden prompt, is sent to the LLM via a POST request as illustrated in Figure. 10. This prompt is not visible to the user and is designed to guide the LLM in generating clinically relevant insights.

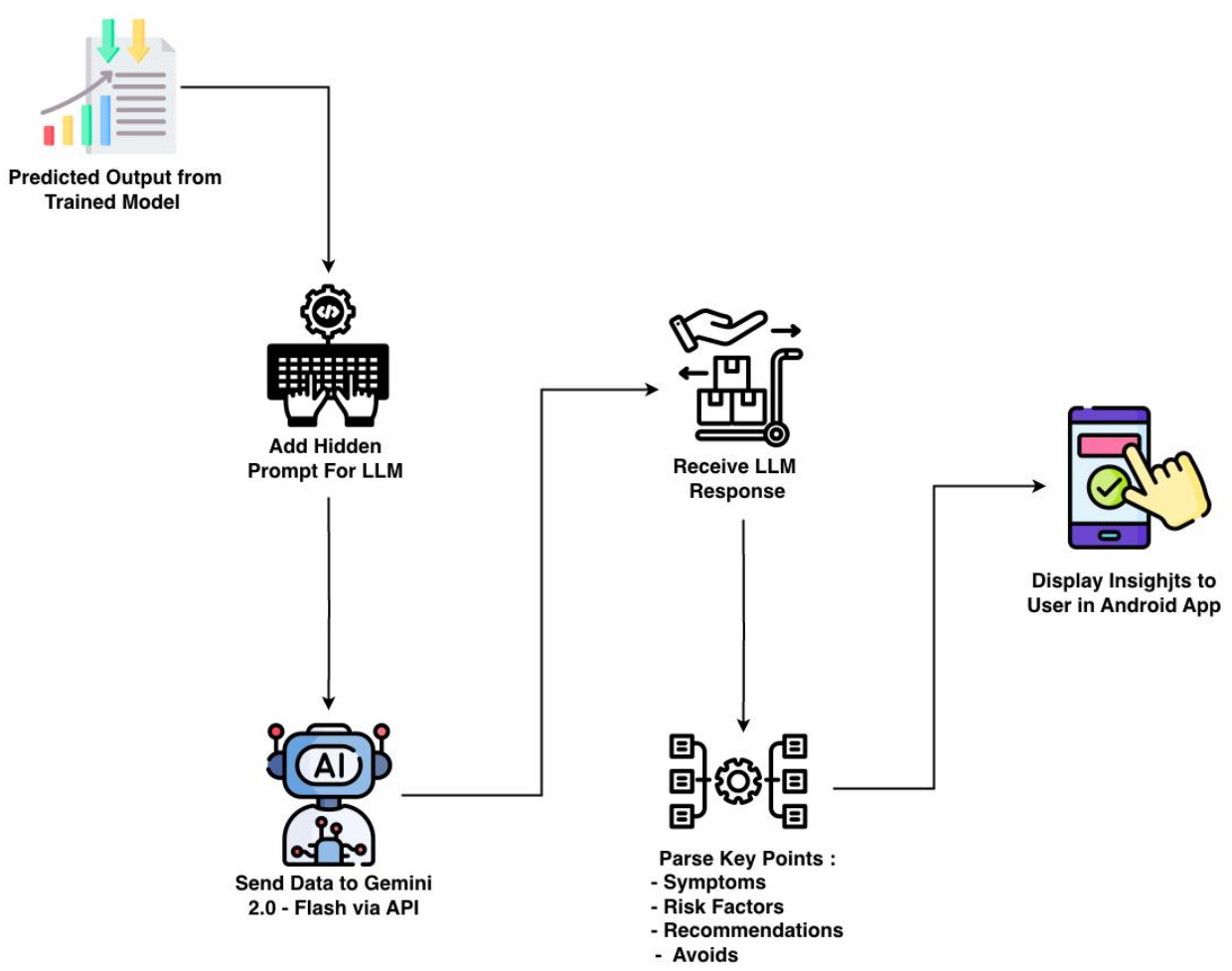

*Fig. 10 LLM interaction workflow for KOA severity insights.*

The LLM processes the input received via a POST request and generates a response that users can access through a GET request. The response is processed by the system to extract valuable information, which shows the estimated knee osteoarthritis grade. The system extracts critical data about essential symptoms together with major risk elements that drive disease development while it offers vital advice about preventive and corrective actions.

The LLM identifies specific actions and specific habits that users must not practice to protect their joints from further damage. The system presents all results in organized formats, which provide users precise instructions for effective management of their knee health.

*2) Visualizing LLM Insights on the Constrained device*

The Android app shows parsed insights to users who can view the complete explanation of the predicted knee osteoarthritis grade together with its symptoms and risks and recommendations and preventive guidance. The complete interaction and results are illustrated in Figure. 11, showing how the LLM enhances the interpretability of the prediction within the device interface.

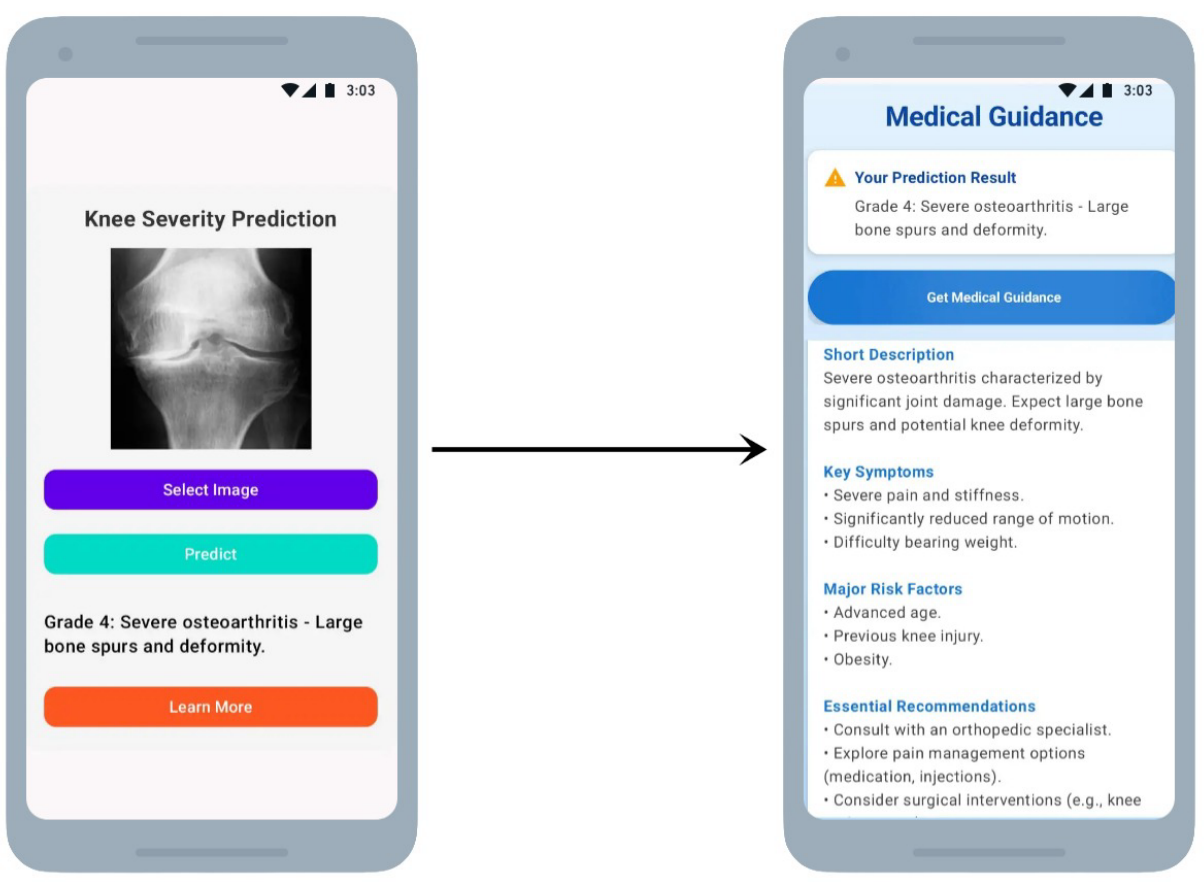

*Fig. 11 Visualization of LLM-generated insights on a constrained mobile device.*

## V. Conclusion

This study successfully develops and analyzes the performance of the designed model on a resource-constrained platform for a prevalent knee problem. It elucidated an AI-driven diagnostic tool that aims to provide a more accessible, robust assessment and evaluation system.

In this study an automated KOA severity classification model using fine-tuned ResNet-18 architecture is developed for the assessment of disease progression. The evaluation of the dataset exhibited a test accuracy of 94.48%, which supports significant potential to determine aspects linked with different grades of KOA.

In order to achieve a low-resource tool, trained model optimization and conversion to TensorFlow Lite are performed. This enables real-time inference on resource-constrained devices without internet connectivity. To generate structured interpretation, LLM is incorporated. This LLM exerts no impact on the prediction process. It simply acts as an informative interface.

Moreover, this study encompasses a clinically challenging KL grade analysis, i.e., grades 1-3. It shows consistent precision, recall, and F1 score values, which outline the model's ability to identify subtle structural variations. Generally, such variations are difficult to address accurately through manual assessments. Overall, this portable assessment tool could assist in early screening, progression, and managing the KOA.

## Acknowledgement

The authors would like to thank the Department of Computer Science & Engineering, Jamia Hamdard, New Delhi, for providing the essential facilities and academic assistance to carry out this research. The publicly accessible Knee Osteoarthritis Dataset and related development tools that made this study possible are also acknowledged by the authors. Additionally, the authors acknowledge Google for granting them access to the Gemini API (Gemini-2.0-flash), which was used in this work to produce interpretative insights.